\documentclass[letterpaper, 10 pt, journal, twoside]{IEEEtran}

\IEEEoverridecommandlockouts                                     

\usepackage{hyperref}
\usepackage{amsmath}
\usepackage{amssymb}
\usepackage{mathptmx} 
\usepackage{mathtools}
\usepackage{xcolor}
\usepackage{algpseudocode}
\usepackage{algorithm} 
\usepackage{graphics} 
\usepackage{graphicx}
\usepackage[font=small]{caption} 
\usepackage{textcomp}
\usepackage[export]{adjustbox}
\usepackage{epstopdf}
\usepackage{subfigure}
\usepackage{float}
\usepackage{epsfig} 
\usepackage{url}
\usepackage{times} 
\usepackage{array}
\newcolumntype{M}[1]{>{\centering\arraybackslash}m{#1}}
\usepackage{booktabs}
\usepackage{bm}

\usepackage{pifont}

\usepackage{listings}
\usepackage{cancel}

\usepackage{soul}

\newcommand{\added}[1]{\textcolor{black}{#1}}
\newcommand{\norm}[1]{\left\lVert#1\right\rVert}

\title{A Shared Autonomy Reconfigurable Control Framework for Telemanipulation of Multi-arm Systems}

\author{Idil Ozdamar$^{1}$, Marco Laghi$^{2}$, Giorgio Grioli$^{1}$, Arash Ajoudani$^{2}$, Manuel G. Catalano$^{1}$, Antonio Bicchi$^{1,3}$
	\thanks{$^{1}$Soft Robotics for Human Cooperation and Rehabilitation (SoftBots), Istituto Italiano di Tecnologia, Genoa, Italy}
	\thanks{$^{2}$Human-Robot Interfaces and Physical Interaction (HRI2), Istituto Italiano di Tecnologia, Genoa, Italy}
	\thanks{$^{3}$Centro di Ricerca ``E. Piaggio'', Universita di Pisa, Largo L. Lazzarino, 1, 56126 Pisa, Italy.}
	\thanks{Manuscript received: February 24, 2022; Revised April 22, 2022; Accepted July 1, 2022.}
	\thanks{This paper was recommended for publication by Editor Angelika Peer upon evaluation of the Associate Editor and Reviewers' comments.} 
	\thanks{\small Correspond to: {\tt\footnotesize idil.ozdamar@iit.it}}
	\thanks{The authors thank Doganay Sirintuna for fruitful discussions, Alberto Justo Lobato for the help during the experiments, and Manuel Barbarossa for the contribution of the experimental setup.}
	\thanks{Digital Object Identifier (DOI): 10.1109/LRA.2022.3191200}
}

\begin{document}

\maketitle

\begin{abstract}
Teleoperation is a widely adopted strategy to control robotic manipulators executing complex tasks that require highly dexterous movements and critical high-level intelligence.
Classical teleoperation schemes are based on either joystick control, or on more intuitive interfaces which map directly the user arm motions into one robot arm's motions. These approaches have limits when the execution of a given task requires reconfigurable multiple robotic arm systems.
Indeed, the simultaneous teleoperation of two or more robot arms could extend the workspace of the manipulation cell, or increase its total payload, or afford other advantages.
In different phases of a reconfigurable multi-arm system, each robot could act as an independent arm, or as one of a pair of cooperating arms, or as one of the fingers of a virtual, large robot hand.
This manuscript proposes a novel telemanipulation framework that enables both the individual and combined control of any number of robotic arms.
Thanks to the designed control architecture, the human operator can intuitively choose the proposed control modalities and the manipulators that make the task convenient to execute through the user interface.
Moreover, through the tele-impedance paradigm, the system can address complex tasks that require physical interaction by letting the robot mimic the arm impedance and position references of the human operator. 
The proposed framework is validated with 8 subjects controlling  4 Franka Emika Panda robots with 7-DoFs to execute a telemanipulation task. Qualitative results of the experiments show us the promising applicability of our framework.
\end{abstract}


\begin{figure}[t]
	\centering
	\includegraphics[width=1\linewidth]{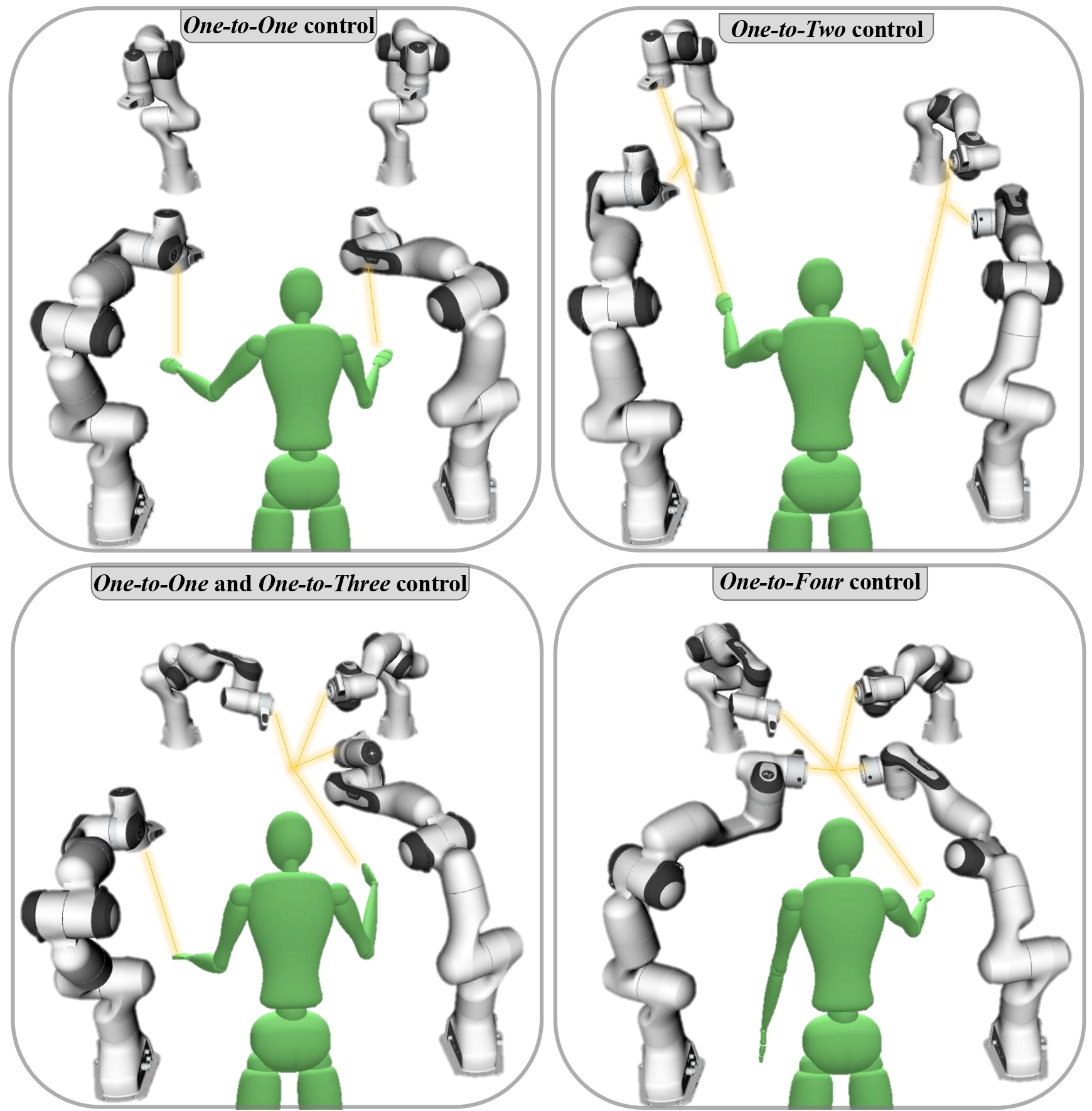}
	\caption{The examples of different possible configurations indicate the user can freely choose and control any number of robots thanks to the reconfigurability of our proposed framework. Light yellow lines show the hands regulating the robots' motion.} \vspace{-4mm}
	\label{fig:cover}
\end{figure}

\vspace{-4mm}
\section{INTRODUCTION}
\label{sec:introduction}

\begin{figure*}[!ht]\vspace{-2mm}
	\centering
	\includegraphics[width=1\linewidth]{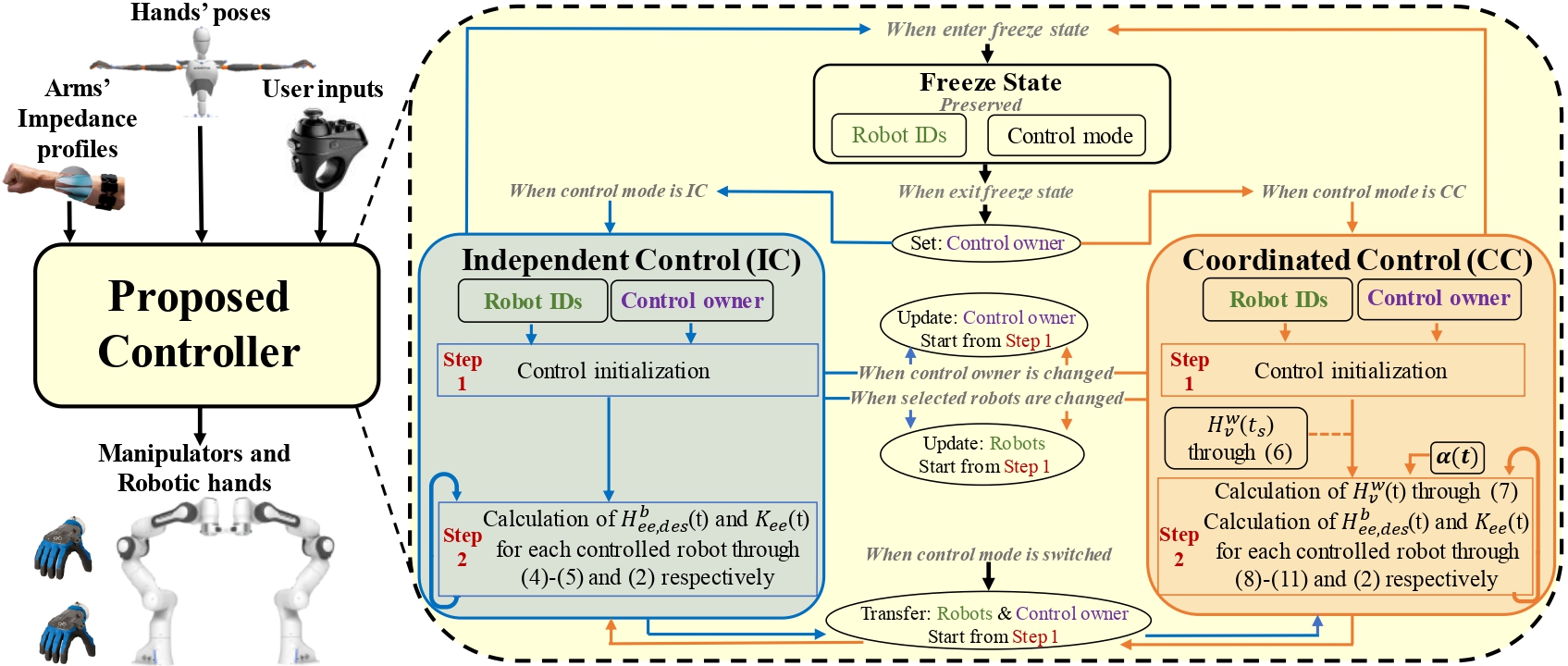}\vspace{-1mm}
	\caption{High level scheme of the proposed control architecture. The reference impedance and the hand poses are sent to the controller together with the current user commands. Then, the end-effectors state (open/close), the desired $H^{b}_{ee,des}(t)$ and $K^{b}_{ee}(t)$ are sent to the robot platform.} \vspace{-5mm}
	\label{fig:control_modes}\vspace{-2mm}
\end{figure*}

\IEEEPARstart{T}{he} essence of teleoperation is captivating owing to integrating human cognitive skills into the system, especially while operating in unknown, hazardous, or dynamic environments.
In this wide researched domain, it is possible to find applications that validate the usability of these systems such as space exploration~\cite{Artigas,Bluethmann,Sheridan}, research and rescue~\cite{Negrello,Hirche}, underwater~\cite{Khatib} and industrial settings~\cite{Shukla}.
In recent years, the technological advancements on the collaborative robot technology and motion capture systems reduced classical limitations of effective teleoperation and renewed the interest in these applications.

In the teleoperation systems, the movements of the slave robot can be controlled by various interfaces such as joystick~\cite{Glynn}, haptic interface~\cite{Martinez,Selvaggio}, 3-D mouse~\cite{Materna,Garate} or motion capture systems~\cite{wearableHaptic,VR,baxterVR,Koenemann}, as in our case.
Among these, it has been found that direct motion mapping is the most intuitive and effective method for the user to operate a robot~\cite{Rakita2}.

The success of telemanipulation also relies on the ability of the system to adapt its interaction with the environment.
In this context, the tele-impedance paradigm~\cite{ajoudani2018reduced} is a powerful tool, as it foresees the estimation of the user arm impedance through surface electromyography (sEMG) and its replication at the robotic counterpart.
In this way the remote manipulator mimics the human dynamic behaviour, increasing safety, adaptability, and efficiency.

With the increase of the complexity of robotic manipulation tasks, the simultaneous use and coordination of multiple robots can be necessary, as they also enlarge the workspace and the range of loads that can be manipulated. In particular, bimanual manipulation becomes increasingly popular~\cite{Edsinger,MAKRIS}, which requires dual-arm robotic systems where the manipulators are coordinated to achieve a mutual goal~\cite{Smith}. To this end, the concept of shared control, in which the robot operates with a degree of autonomy to reduce user effort, has been introduced to teleoperation systems where the human and the robot have a common mission. To benefit from shared control in multi-arm systems, Rakita et al.~\cite{Rakita} used bi-manual motion vocabulary to blend user translation and rotation control inputs with known translation and rotation paths. In~\cite{Sun}, the contact forces and object orientation are self-regulated so that the operator only needs to handle the position of the 3-DoF master haptic device. Lin et al.~\cite{Lin} adopted a vision-based object detection method to automate the arms' grasping action, but limited to a single hand. In~\cite{Amanhoud}, the authors are interested in tasks requiring two people, hence they complement the human arms with two robotic arms controlled via a bi-pedal foot interface. In our previous work~\cite{laghi2018shared}, we showed that supporting and moving a load with the synergistic coordination of robots was particularly suitable and effective for symmetric tasks. All the studies mentioned above implement different shared control telemanipulation strategies, but are always confined to two arm telemanipulation.

In the literature, very few studies focus on employing more than two manipulators in teleoperation. \cite{Tung} collects demonstrations of multi-arm tasks for use in the Imitation Learning paradigm with three robotic arms, but controlled via smartphone by separate users at different locations (one user for each arm). In~\cite{multiarm}, the authors consider multi-arm object manipulation scenario using either three or four arms, and introduce a method to increase the manipulability index of each arm. Nonetheless, the object was already grasped by the robots, and the work does not cover the situations of approach, grasp, and release of the object. In addition, both works validations are limited to simulation only.

In this work, we develop a novel telemanipulation framework that extends the previous proposed in \cite{laghi2018shared}. The new solution is designed to overcome the limitation of two robots, their configuration constraint (when one arm controls the end-effectors, they always face each other), and handle any number of manipulators. In particular, the new introduced features are:

\begin{itemize}
    \item generalization of the coordinated control mode to allow any possible relative pose between the end-effectors;
    \item extension of both independent and coordinated control modes to control any number of robots with one arm;
    \item relinquishment and retainment of the preferred robots' control when it is requested;
    \item creation of a resources (robots) manager which enables the user to easily select and reconfigure the controlled manipulators and the control modalities on-the-fly.
\end{itemize}
Through such features, the user can benefit from the high dexterity of one-to-one coupling as well as the combined control of multiple robots with the integration of shared autonomy into the framework. The tele-impedance~\cite{ajoudani2018reduced} control paradigm is also adopted in the interest of safety and interaction performance of the robots with their environment. Besides the technical aspects of teleoperation, intuitiveness, attention needs, and comfort are taken into account while implementing the reconfigurability structure to ensure users’ acceptability.

The rest of this article is organized as follows: Section II presents the implementation of our proposed multi-arm shared control telemanipulation framework. It covers the method of robot arm control, the developed control modalities, and explains how the user reconfigures the system through the user interface. Section III describes the experimental setup, procedure, and the conducted user study. Finally, in Sections IV and V, we discuss the results and conclude the study with possible future research directions.

\section{PROPOSED FRAMEWORK}
\label{sec:methodology}
The main contribution of this study is the definition of a framework in which any number of robots can be simultaneously controlled through teleoperation.
This framework enables users to create, modify and teleoperate different subgroups of arms in accordance to the task to be performed.

\begin{figure}\vspace{-2mm}
	\centering
	\includegraphics[width=.87\linewidth]{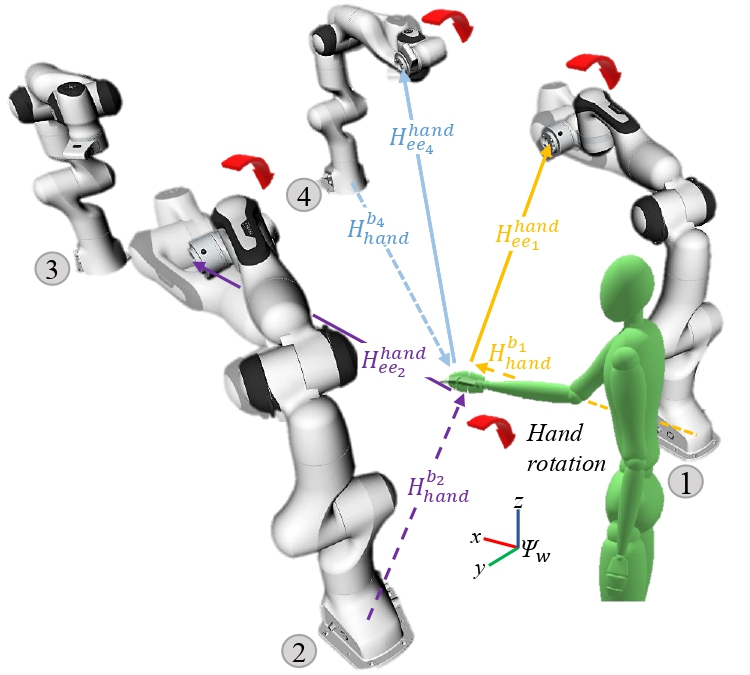}\vspace{-2mm}
	\caption{The hand and end-effector frames' mapping scheme when the user controls 3 robots with Independent Control strategy. The numbers located near the robot bases represent their IDs.} \vspace{-6mm}
	\label{fig:indep_human_robot}
\end{figure}

\vspace{-3mm}
\subsection{Control of a Single Robotic Arm}
\label{subsec:robot_control}

In this work, the tele-impedance control \cite{ajoudani2018reduced} is used to regulate the motion and the dynamic behavior of each robot. To do so, user's arm stiffness index $s_a(t)$ and desired end-effector pose are sent as references to the robot controller. The desired pose frame $\Psi_{ee,des}$ w.r.t. the robot base frame $\Psi_b$ is expressed through $H^{b}_{ee,des}(t)$, with $H^{j}_{i}(t)$ being the homogeneous transformation matrix of the frame $\varPsi_i$ w.r.t. $\varPsi_j$ at time instant $t$ defined as 
\begin{equation}\vspace{-2mm}
    H^{j}_{i}(t) =\begin{bmatrix}
	R^{j}_{i}(t) & l^{j}_{i}(t)\\
	0_{(1\times 3)} & 1
	\end{bmatrix}_{(4\times 4)},  
\end{equation}
with $R^{j}_{i} \in \mathbb{R}^{3\times 3}$ and $l^{j}_{i}\in \mathbb{R}^{3}$ being the rotation matrix and the position vector, respectively.

The desired stiffness matrix at the end effector sent to the robot controllers are defined as follows

\begin{equation}\vspace{-2mm}
	K_{}(t) =
	\begin{bmatrix}
	k_{l}(t) I_{3 \times 3} & 0_{3\times 3}\\
	0_{3\times 3} & k_{\omega}(t) I_{3 \times 3}
	\end{bmatrix},
	\label{eq:stiffM}
\end{equation}
where $k_{l}(t)$ and $k_{\omega}(t)$ represents the desired unidimensional linear and rotational stiffnesses, respectively, set  proportionally to the user's stiffness index $s_a(t)$ with the same procedure suggested in \cite{laghi2018shared}.

The desired joint torques $\tau_{des}(t)$ are therefore calculated by a variable Cartesian impedance controller as
\begin{equation}\vspace{-2mm}
\begin{aligned}
	\tau_{des}(t) = &J^T(q(t)) (K(t)\tilde{x}(t) + D(t)\dot{\tilde{x}}(t))+\\ &\added{N(q(t))(K_q\tilde{q}(t) + D_q\dot{\tilde{q}}(t))+}\\
	&M(q(t))\ddot{q}(t) + C(q(t),\dot{q}(t))\dot{q}(t)+g(q(t)),
\end{aligned}
\label{eq:impedance_control}
\end{equation}
where $q(t)$ is the joint position vector, $J(q(t))$ is the robotic arm Jacobian, $\tilde{x}(t)$ is the position and orientation error vector between desired pose, $H^b_{ee,des}(t)$, and the actual pose, $H^b_{ee}(q(t))$, and $\dot{\tilde{x}}(t)$ is its time derivative. $K(t)$ and $D(t)$ are the variable stiffness and damping matrices. $D(t)$ is the damping matrix at the end-effector, that can be defined through a relation with $K(t)$. In this work, the damping factor $\xi$ is set to 1. $M(q(t))$ is the joint mass matrix, $C(q(t),\dot{q}(t))$ is the centrifugal/Coriolis term, and $g(q(t))$ represents the gravity compensation torque contributions.The term $N(q(t))(K_q\tilde{q}(t) + D_q\dot{\tilde{q}}(t))$ is the nullspace impedance used to control the robot redundancy.
$\tilde{q}(t) = q_d(t) - q(t)$ and $\dot{\tilde{q}}(t)$ are the joint position and velocity error, respectively.
$K_q$ and $D_q$ are the joint space stiffness and damping matrices.
In this work, $q_d$ is constantly set to $0$, to force the joints to stay away from their position limits.
This can naturally be changed on the need, e.g. to optimize the internal robot configuration w.r.t. the task purpose or to avoid collisions with the environment.
$N(q(t))$ is the null space projector, here set as $N(q(t))=I-J^T(q(t)){J^\dagger}^T(q(t))$, with $\cdot^\dagger$ being the pseudo-inverse operator. Other projectors can be used, if different performances are foreseen (e.g., see \cite{dietrich2015overview}).

\begin{figure}\vspace{-2mm}
	\centering
	\includegraphics[width=0.92\linewidth]{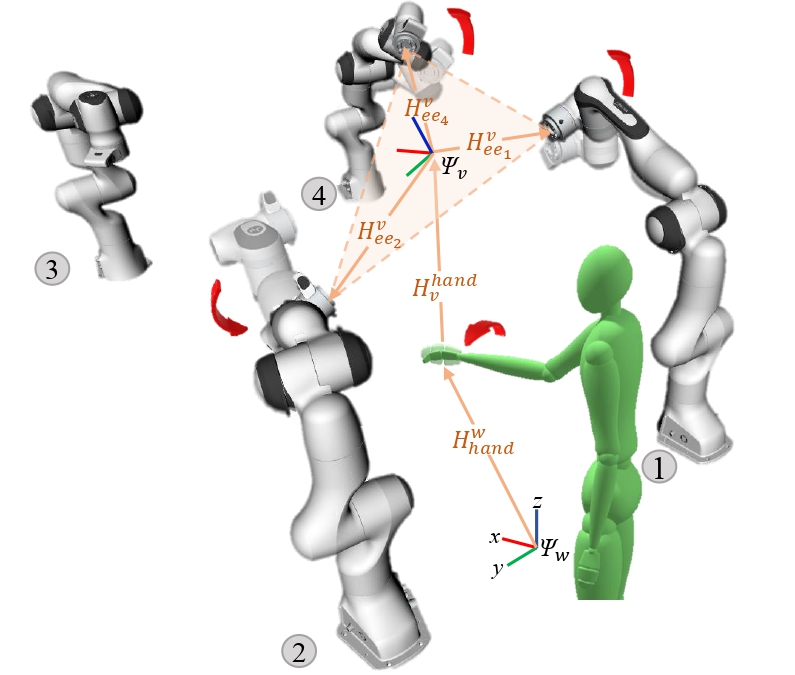}\vspace{-2mm}
	\caption{The hand, virtual, and end-effector frames' mapping scheme when the user controls 3 robots with Coordinated Control strategy. The numbers located near the robot bases represent their IDs.}\vspace{-6mm}
	\label{fig:coord_human_robot}
\end{figure}

\vspace{-3mm}
\subsection{Control Modalities for Multiple Robotic Arms}
\label{subsec:control_strategies}

The proposed control architecture comprises three main modalities, namely \textit{Coordinated Control (CC)}, \textit{Independent Control (IC)}, and  \textit{Freeze State}. The scheme of such architecture and the relationship between control modalities can be seen in Fig.~\ref{fig:control_modes}. The working principles of these units and the protocol applied in the transitions between each other are designed to allow the user to rule the multi-manipulator system intuitively.
Each proposed control strategy requires two inputs which are obtained through the joysticks:
\begin{itemize}
    \item \textit{Robots IDs}: This input indicates which robots are selected by the user to be controlled. When the user selects $n \ge 2$ number of robots, a robot group is created. The robots within a group will be moved together through one of the control strategies. 
    \item \textit{Control owner}: It defines the user hand selected to teleoperate the robots group. In the following sections, $\varPsi_{hand}$ is used to indicate the frame of the hand which is the \textit{Control owner} of the selected strategy.
\end{itemize}

During telemanipulation, $H^b_{ee,des}(t)$ and $K_{ee}(t)$ are calculated individually for each selected robot by the proposed control strategies. Then, the local controller of each robot calculates the desired joint torques through Eq. \eqref{eq:impedance_control}.

\subsubsection{Independent Control (IC)}\label{subsubsec:independent_control_mode}~\

The main goal of this strategy is to mimic the human hands' motion by any number of robots' end-effectors.
When the control starts at time $t=t_s$, its initialization (Step 1, see  Fig.~\ref{fig:control_modes}) is performed in order to obtain the initial relative rotation and translation between the hand frame and the end-effector frames of the selected robots: $R^{hand}_{ee_i}(t_s)$ and ${^{b_i}l^{hand}_{ee_i}}(t_s)$ between the hand and the end-effectors' are obtained from $H^{b_i}_{hand}(t_s)$ and $H^{hand}_{ee_i}(t_s)$ for each robot (see Fig.~\ref{fig:indep_human_robot}).
Note that the translation vector of the hand w.r.t. end-effector is expressed in the robots' base frame, so defined as
\begin{align}\vspace{-1mm}
  {^{b_i}l^{hand}_{ee_i}}(t_s) =
  R^{b_i}_{hand}(t_s){l^{hand}_{ee_i}}(t_s),
\end{align} 
where $\varPsi_{b_i}$ and $\varPsi_{ee_i}$ represent the base and the end-effector frames of the $i$-th robot. 

After the initialization, the control loop (Step 2) starts its execution. The desired pose of each robot is calculated as
\begin{align}\vspace{-1mm}
    H^{b_i}_{ee_i,des}(t) = 
    \begin{bmatrix}
    R^{b_i}_{hand}(t)R^{hand}_{ee_i}(t_s) & l^{b_i}_{hand}(t)+{}^{b_i}_{}l^{hand}_{ee_i}(t_s)\\
    {0}_{1\times 3} & 1
    \end{bmatrix}.
    \label{eq:independent_Hmatrix}
\end{align}
In \eqref{eq:independent_Hmatrix}, the current hand rotation $R^{b_i}_{hand}(t)$ is multiplied by initial rotational offset $R^{hand}_{ee_i}(t_s)$, that comes from Step 1. Similarly, the translational offset, ${^{b_i}l^{hand}_{ee_i}}(t_s)$, is added to the current base to hand position vector, $l^{b_i}_{hand}(t)$. In this way, hands' rotation, and translation can be directly transferred to the robot ends. Furthermore, the robots' stiffness is regulated through the mapping method in \cite{laghi2018shared}, to mimic the stiffness profile of the user arm commanding the robots.

\subsubsection{Coordinated Control (CC)}\label{subsubsec:coordinated_control_mode}~\

Within this control strategy, the user manipulates the translation and the rotation of the virtual frame that lies between the robots' end-effectors (see  Fig.~\ref{fig:coord_human_robot}).
In Step 1, when the control strategy is initialized at $t=t_s$, this virtual frame $\Psi_v$ is created at the midpoint between robots' end-effectors and oriented as the command hand:
\begin{equation}\vspace{-1mm}
\begin{aligned}
    H^{w}_{v}(t_s) &= 
    \begin{bmatrix}
    R^{w}_{hand}(t_s) & l^{w}_{v}(t_s)\\
    0_{1\times 3} & 1
    \end{bmatrix};\\
    l^{w}_{v}(t_s) &= \frac{\sum_{i}^{}l^{w}_{ee_i}(t_s)}{n}, 
\end{aligned}
\label{eq:virtual_frame}
\end{equation}
where $\varPsi_w$, $\varPsi_v$, $\varPsi_{hand}$, $\varPsi_{ee_i}$, and $n$ are the world frame, virtual frame, hand frame, end-effector frame of the $i$-th robot, and the total number of robots controlled in this mode, respectively.
$\varPsi_v$ is then linked to $\varPsi_{hand}$ with the technique described in the IC subsection: $R^{hand}_{v}(t_s)$ and ${^{w}l^{hand}_{v}}(t_s)$ are registered and then used to calculate the virtual frame pose during the control loop (Step 2) as
\begin{align}\vspace{-1mm}
    H^{w}_{v}(t) = 
    \begin{bmatrix}
    R^{w}_{hand}(t)R^{hand}_{v}(t_s) & l^{w}_{hand}(t)+{^{w}l^{hand}_{v}}(t_s)\\
    {0}_{1\times 3} & 1
    \end{bmatrix}.
\end{align}

With a strategy similar to the one presented in \cite{laghi2018shared}, the robot end-effectors are linked to the virtual frame through a virtual prismatic joint.
Given $H^{v}_{ee_i}(t_s)$, the desired pose of the $i$-th end effector w.r.t. the virtual frame is calculated as
\begin{equation}\vspace{-2mm}
    H^{v}_{ee_i,des}(t) = 
    \begin{bmatrix}
    R^{v}_{ee_i,des}(t_s) & {l}^{v}_{ee_i,des}(t)\\
    {0}_{1\times 3} & 1
    \end{bmatrix},
    \label{eq:eedes_wrt_virtual_frame}
\end{equation}
with
\begin{equation}\vspace{-1mm}
	{l}^{v}_{ee_i}(t) = \begin{cases}
	{l}^{v}_{ee_i}(t-1) &\textrm{if} \ \norm{{l}^{v}_{ee_i}(t_s)\alpha(t)}<{l}_{min}\\
	{l}^{v}_{ee_i}(t-1) &\textrm{if} \ \norm{{l}^{v}_{ee_i}(t_s)\alpha(t)}>{l}_{max}\\
	{l}^{v}_{ee_i}(t_s)\alpha(t) \quad &\textrm{otherwise}
	\end{cases} \ \forall t\ge t_s,
	\label{eq:closure}
\end{equation}
saturated between lower, ${l}_{min}$, and upper, ${l}_{max}$, limits to prevent the robots from colliding with each other or command unreachable positions. In \eqref{eq:closure} $\alpha(t) = 1+ \int_{t_s}^{t}\epsilon(t)dt$ and $\epsilon(t)$ is utilized to increase or decrease the $\alpha(t)$ according to the joystick command. The $\epsilon(t)$ defined as
\begin{equation}
	\epsilon(t) = \begin{cases}
	\epsilon &\textrm{if joystick command is increase},\\
	-\epsilon &\textrm{if joystick command is decrease},\\
	0 \quad &\textrm{otherwise}.
	\end{cases}
	\label{eq:closure_vel}
\end{equation}
Finally, the pose sent as reference to the $i$-th robot controller is
\begin{equation}\vspace{-1mm}
    H^{b_i}_{ee_i,des}(t) = H_{w}^{b_i}H^{w}_{v}(t)H^{v}_{ee_i,des}(t) = H^{b_i}_{v}(t)H^{v}_{ee_i,des}(t),
\label{eq:des_pose_coord}
\end{equation}
where the current base to virtual frame transformation of the $i^{th}$ robot, $H^{b_i}_{v}(t)$, multiplies the desired transformation matrix of the robot's end-effector desired frame w.r.t.virtual frame, $H^{v}_{ee_i,des}$ (see Fig.~\ref{fig:coord_human_robot}).

Comparing the above equations, and in particular \eqref{eq:eedes_wrt_virtual_frame}, with the ones presented in \cite{laghi2018shared}, it is possible to notice that the main difference resides in the orientation commanded to the robot.
Indeed, while in the synergistic control of \cite{laghi2018shared} the robots were forced to face each other, here their relative orientation is kept equal to the one at $t_s$. This trick allows the user to freely choose the relative orientation between the end-effectors, enabling the execution of a larger variety of coordinated tasks. Please see the video footage \footnotemark\footnotetext{\label{video} The video can be found at: \url{https://youtu.be/zef-86tK8xU}} for the demonstrations of the control modalities.

\subsubsection{Freeze State}\label{subsubsec:freeze_state}~\

When dealing with multiple manipulators, one can take advantage of keeping the robots at a definite position with the \textit{Freeze State} proposed in this study. If the user chooses to enter into the \textit{Freeze State}, the robots' reference poses and stiffnesses remain constant regardless of user motion, and the \textit{Robot IDs} and the \textit{Control owner} information are preserved inside this state (see Fig.~\ref{fig:control_modes}). Then, this information is used when exiting from this state, in order to continue telemanipulation with the same robots and the same control strategy.

\begin{figure}[!t]\vspace{-1mm}
	\centering
	\subfigure{
		\includegraphics[width = 0.89\linewidth,trim={0 0cm 0 0cm},clip]{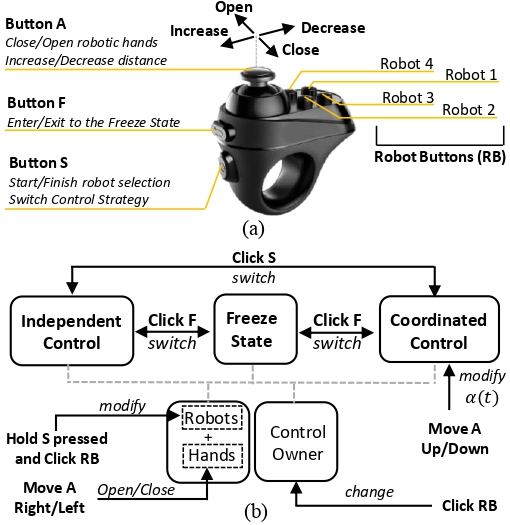}
		\label{fig:exp_res}} 
	\subfigure{
		\includegraphics[width = 1\linewidth]{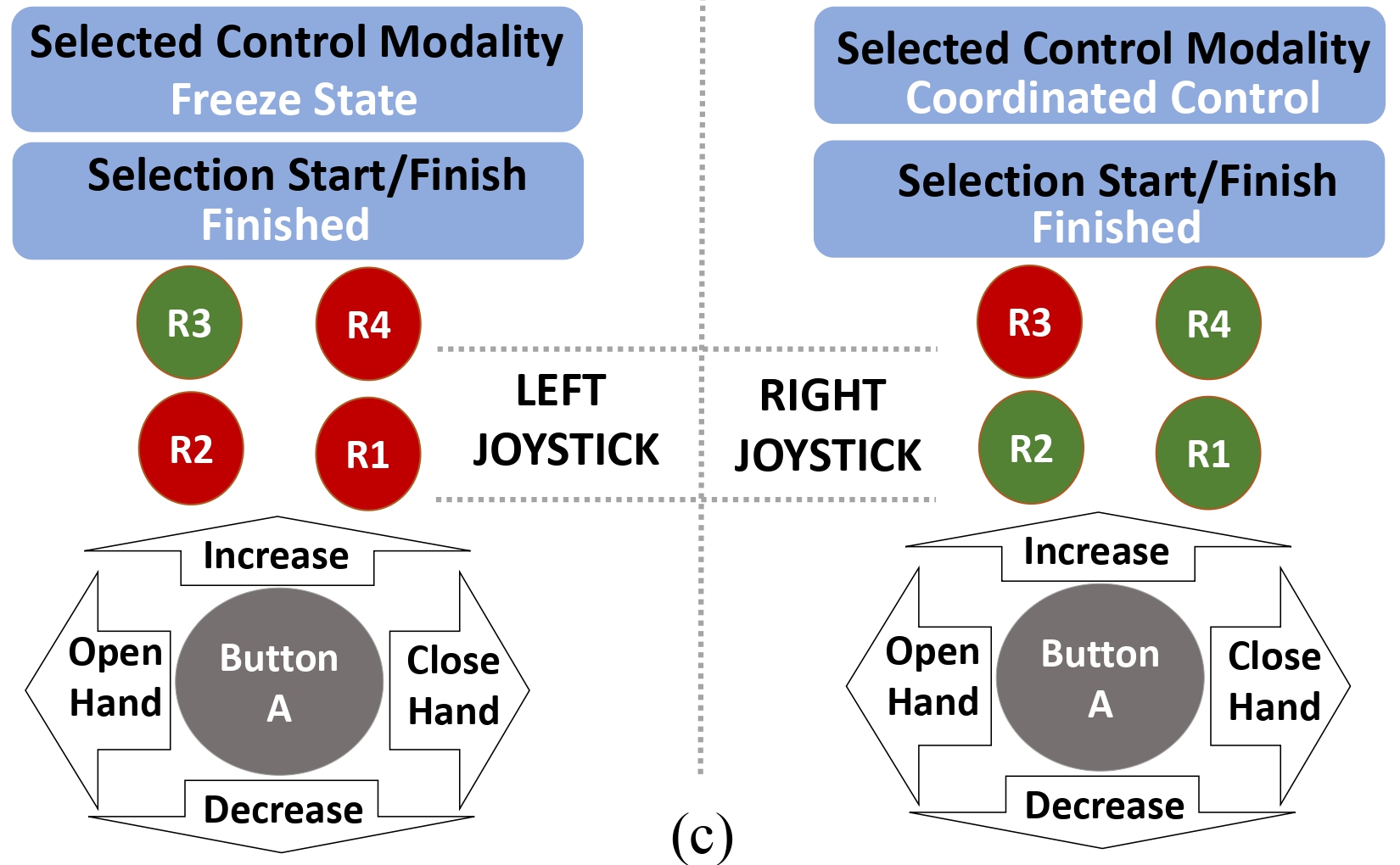}
		\label{fig:gui}}\vspace{-2mm}
	\caption{(a) The user interface, (b) the functionality of the buttons in the control architecture and (c) the graphical user interface (GUI).}
	\label{fig:user_interface}\vspace{-8mm}
\end{figure}

\vspace{-3mm}
\subsection{User Interface}
\label{subsec:interface}

In order to communicate with the system, the operator is provided with two joysticks, one per hand. Fig.~\ref{fig:user_interface}(a) shows the used joystick and the functionality of each button, and Fig.~\ref{fig:user_interface}(b) depicts how the proposed control framework is managed by the user interface commands.
The robots are selected by clicking the corresponding RB button while keeping the button S pressed until the selection ends.
Whichever hand that makes the selection, this hand becomes the \textit{Control owner} of the selected robots. If only one robot is chosen, its control strategy is assigned to the independent control by default. When multiple robots are selected ($n \ge 2$), a robot group is created so that they can move in tandem. The user can modify a robot group by holding the button S pressed, and then clicking the RB buttons relative to the robot to be added/removed.

In order to switch the \textit{Control owner} of the whole robot group, it is sufficient to click the RB button that corresponds to any robot within the group, without pressing the button S.
The hand that makes the selection will be the new \textit{Control owner} of this robot group. If this hand is already in charge of other robots' control, these robots are combined with transferred robots to form a single group. To change the control strategy of a robot group, the user has to click the button S. Control strategies (IC and CC) toggle between each other with each click of button S. When the user clicks the button "F" the same toggle rule is valid for freezing or unfreezing the robot. 

In addition, a graphical user interface (GUI) has been placed in front of the users so that they can see the button functions and track their current selections.
Fig.~\ref{fig:user_interface}(c) shows the developed GUI in a time instant in which the third robot (R3) is in \textit{Freeze state} and assigned to the left user hand (left side), while robots R1, R2, and R4 are active and controlled in \textit{Coordinated Mode} by the right user hand (right side). On the other hand, if it is desired to use more robots in the system, the joystick can be replaced by another alternative because the proposed framework does not depend on the hardware of the interface.
\vspace{-3mm}

\section{EXPERIMENTS}
\label{sec:experiments}

To adequately evaluate the performance of the proposed telemanipulation framework, we conducted a multi-subject experimental campaign where participants were asked to execute a task in which they need to control the robots through the different strategies. 
In this section, we describe the experimental setup, the campaign, and report the experiment results.

\vspace{-3mm}
\subsection{Experimental Setup}
\label{subsec:exp_setup}
In this study, the proposed telemanipulation framework has been validated using four Panda robotic arms by Franka Emika\footnote{Franka Emika: \url{https://www.franka.de/}}.
As illustrated in Fig.~\ref{fig:cover}, the robots were placed so to form a $0.8\times 1.9$ m rectangle.
Each robot is provided with a Pisa/IIT SoftHand\cite{della2017quest} as an end-effector, to permit the user to easily grasp and handle objects/tools at the remote side.

In order to capture the movement of human operator hands, an Xsens MVN\footnote{Xsens: \url{https://www.xsens.com/}} suit was used during the experiments.
Owing to this motion capture system, the real-time measurements of human hand positions were obtained with high precision.
In addition, two wearable devices equipped with eight surface electromyographic (sEMG) electrodes (MYO Armband) were placed on the forearms of the participants and used to extrapolate the user's arm stiffness $s_a(t)$. This is then mapped to
the desired unidimensional stiffnesses, $k_{l}(t)$ and $k_{w}(t)$ in \eqref{eq:stiffM}, which are in turn saturated between $k_{l,min} = 100$~N/m, $k_{\omega,min} = 10$~Nm/rad, $k_{l,max} = 600$~N/m, $k_{\omega,max} = 60$~Nm/rad, taking in consideration the robots capabilities. A detailed description of this procedure can be found in \cite{laghi2018shared}. The null space stiffness of \eqref{eq:impedance_control} was set $K_q=I_{7\times7}5$~Nm/rad, and the null space damping $D_q$ was calculated with $\xi = 1$.

For the interface described in Sec.~\ref{subsec:interface}, we utilized two Magicsee R1 joysticks. The whole control framework was created through the Robot Operating System (ROS)\footnote{ROS: \url{https://www.ros.org/}}, which provides the necessary communication and visualisation utilities.

\begin{figure*}[ht]\vspace{0mm}
	\centering
	\subfigure {\includegraphics[width = 0.247\linewidth,trim={2.1cm 0cm 2cm 0.5cm},clip]{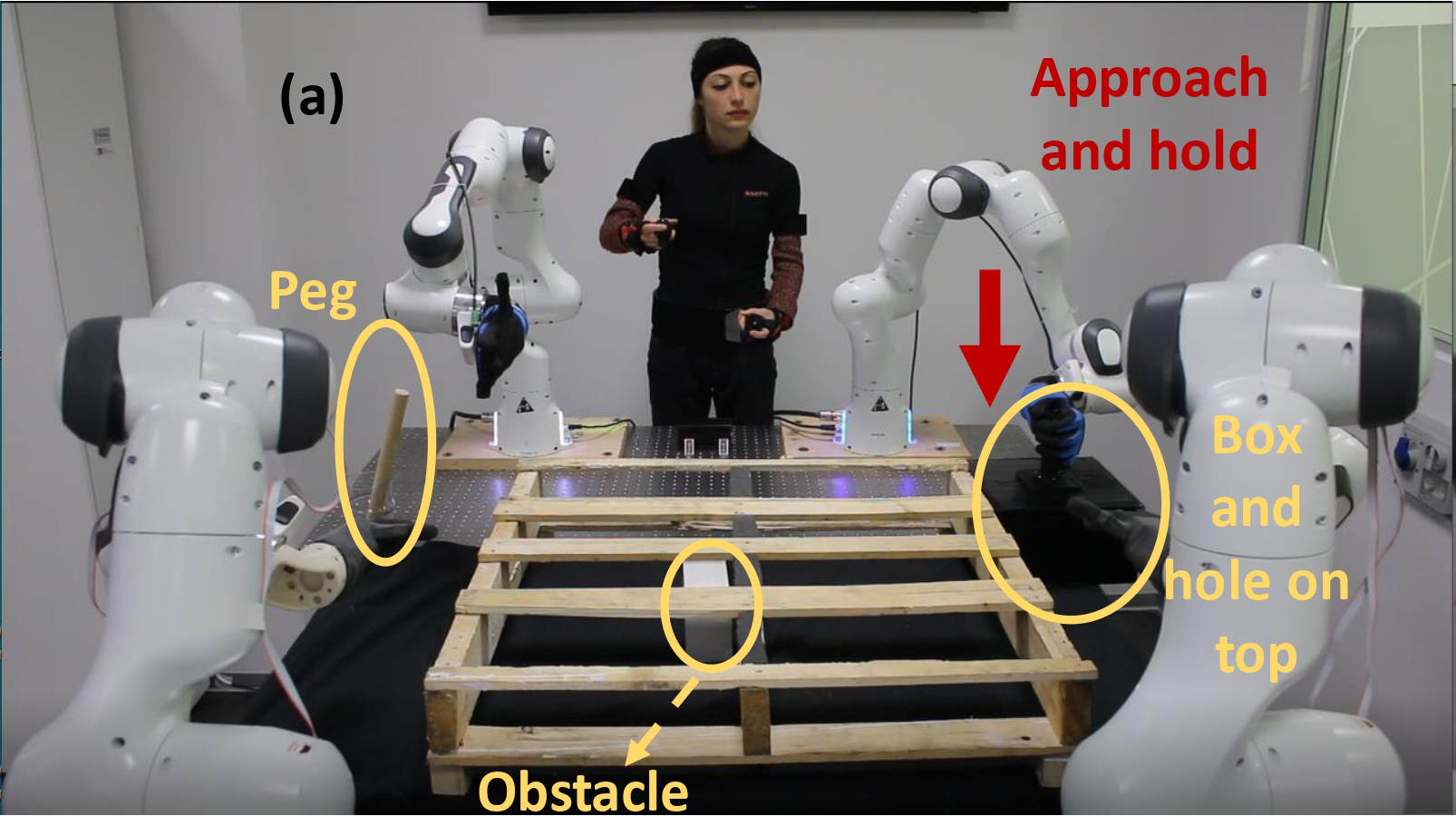}
		\label{fig:task_frame_2}}\hspace{-0.2cm}
	\subfigure {\includegraphics[width = 0.247\linewidth,trim={2.1cm 0cm 2cm 0.5cm},clip]{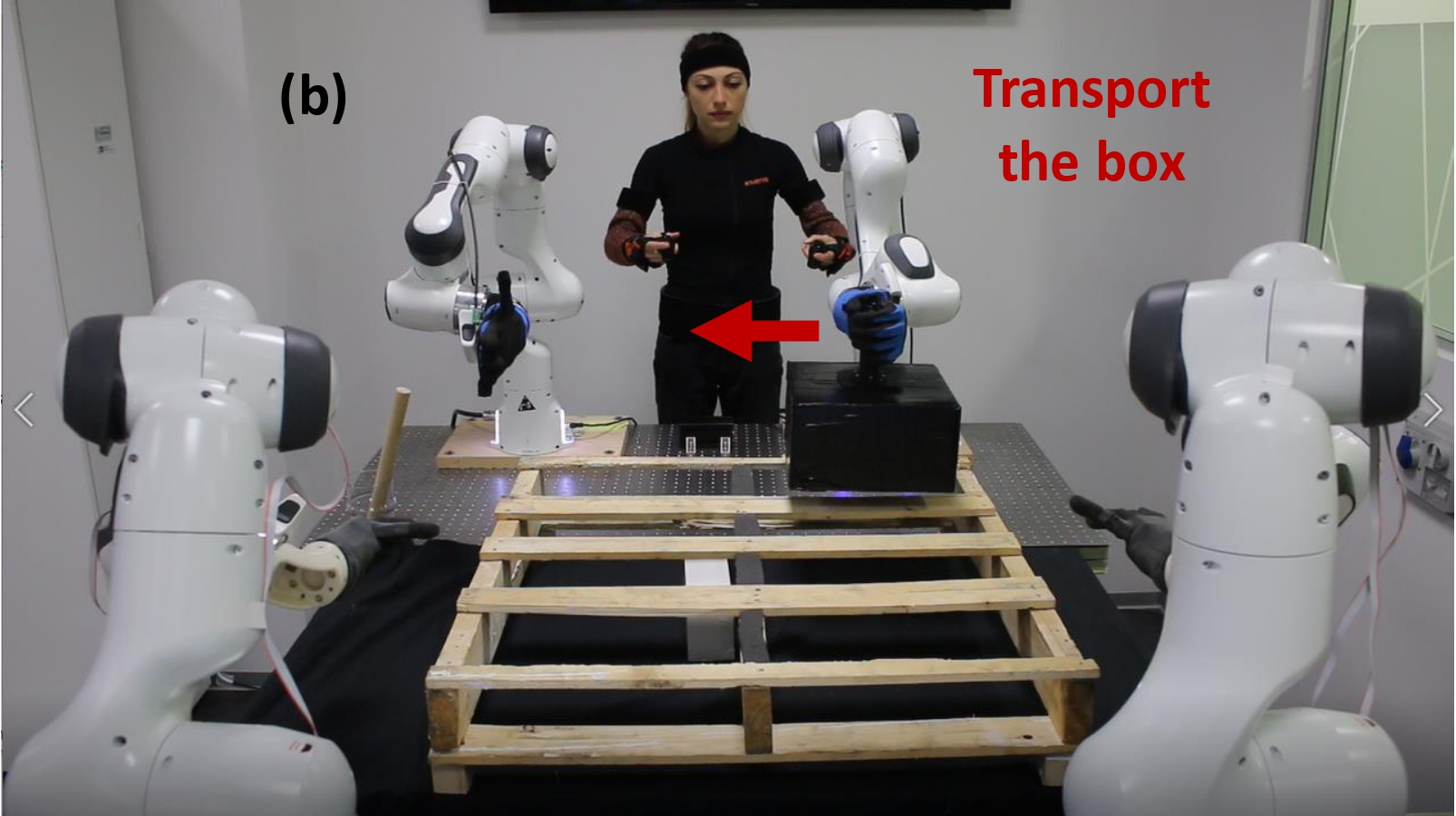}
		\label{fig:task_frame_3}}\hspace{-0.2cm}
	\subfigure {\includegraphics[width = 0.247\linewidth,trim={2.1cm 0cm 2cm 0.5cm},clip]{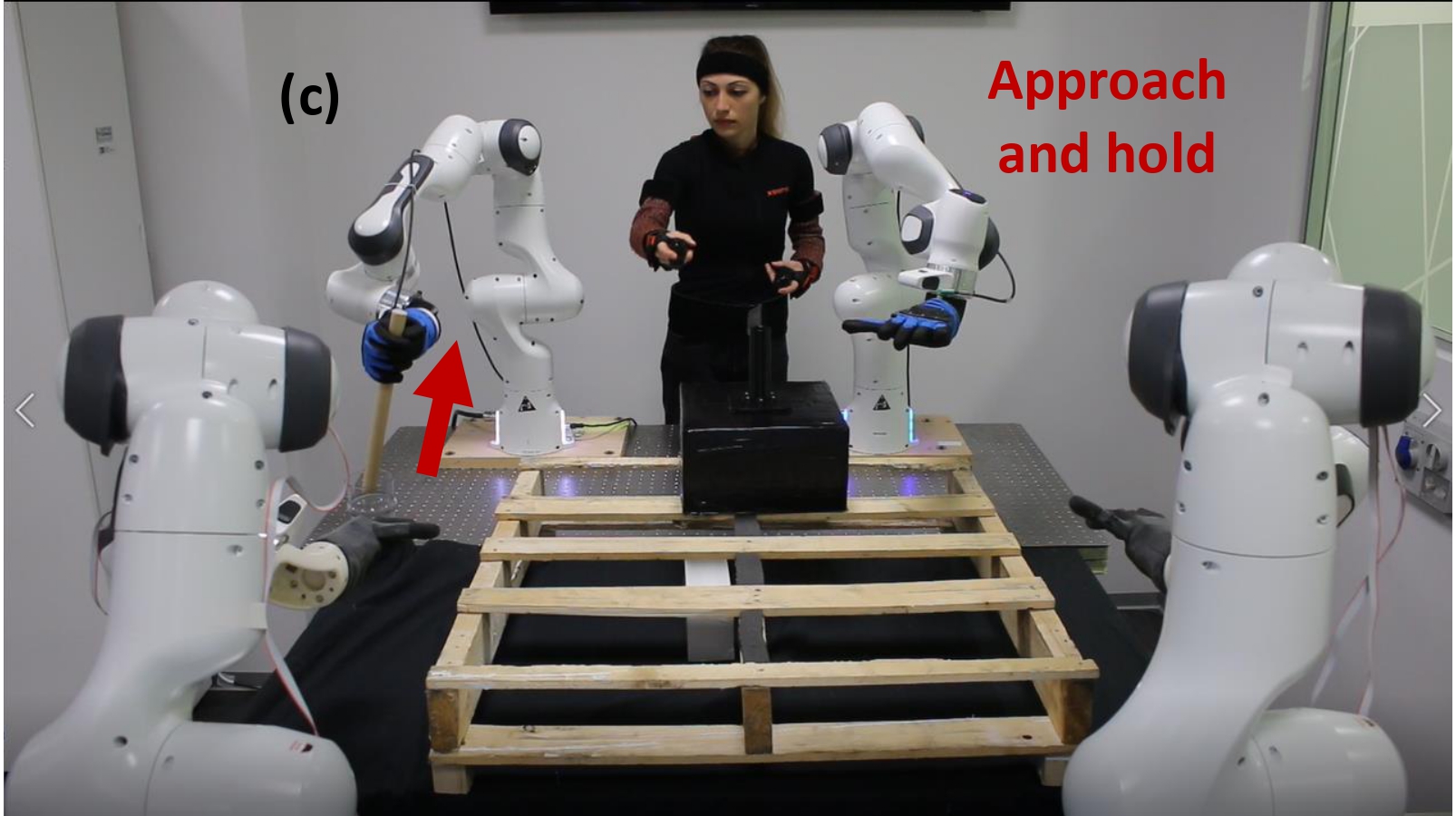}
		\label{fig:task_frame_4}}\hspace{-0.2cm}
	\subfigure {\includegraphics[width = 0.247\linewidth,trim={2.1cm 0cm 2cm 0.5cm},clip]{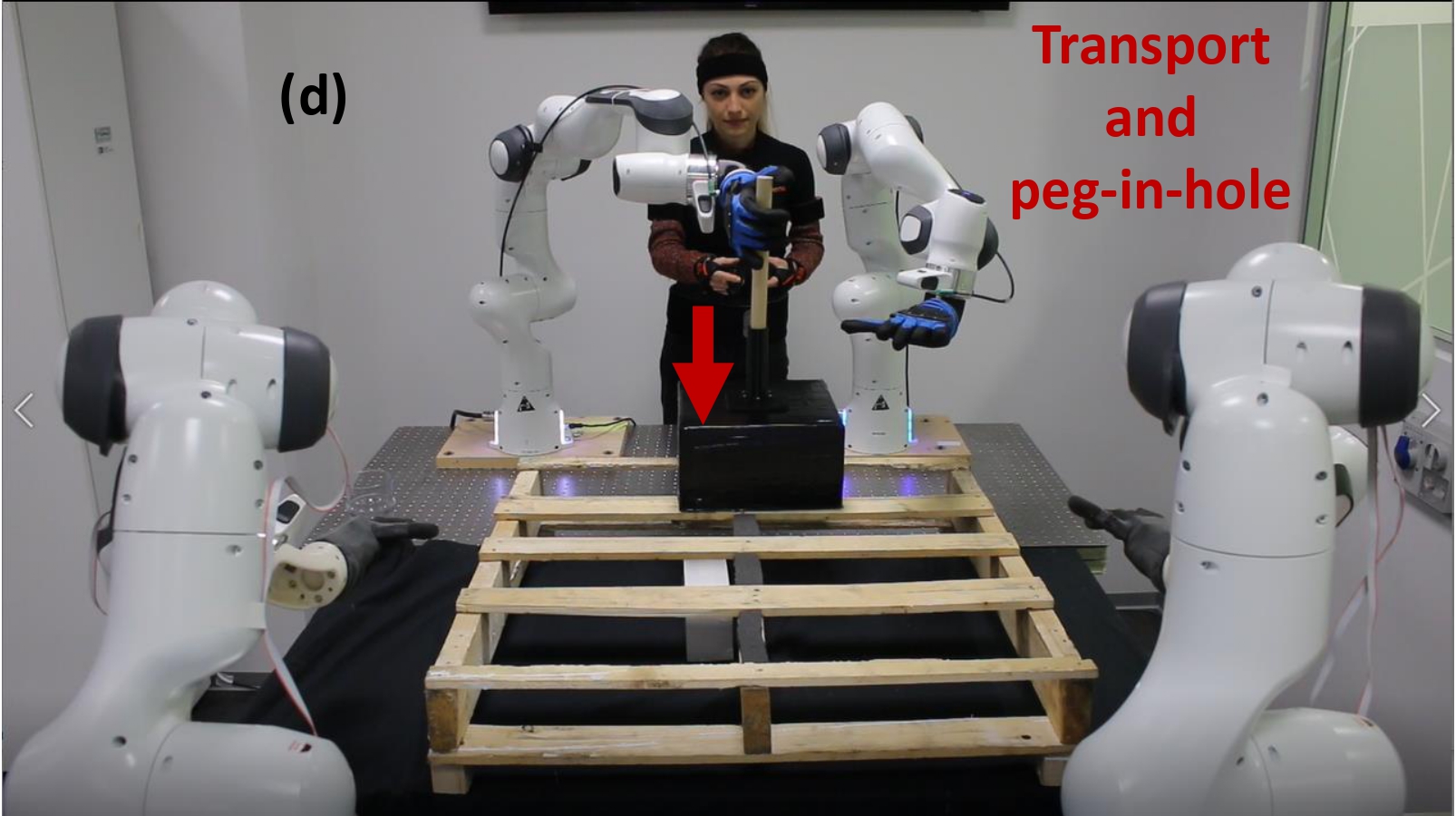}
		\label{fig:task_frame_5}}\hspace{-0.2cm}
		\vspace{-0.3cm}

	\subfigure{\includegraphics[width = 0.247\linewidth,trim={2.1cm 0cm 2cm 0.5cm},clip]{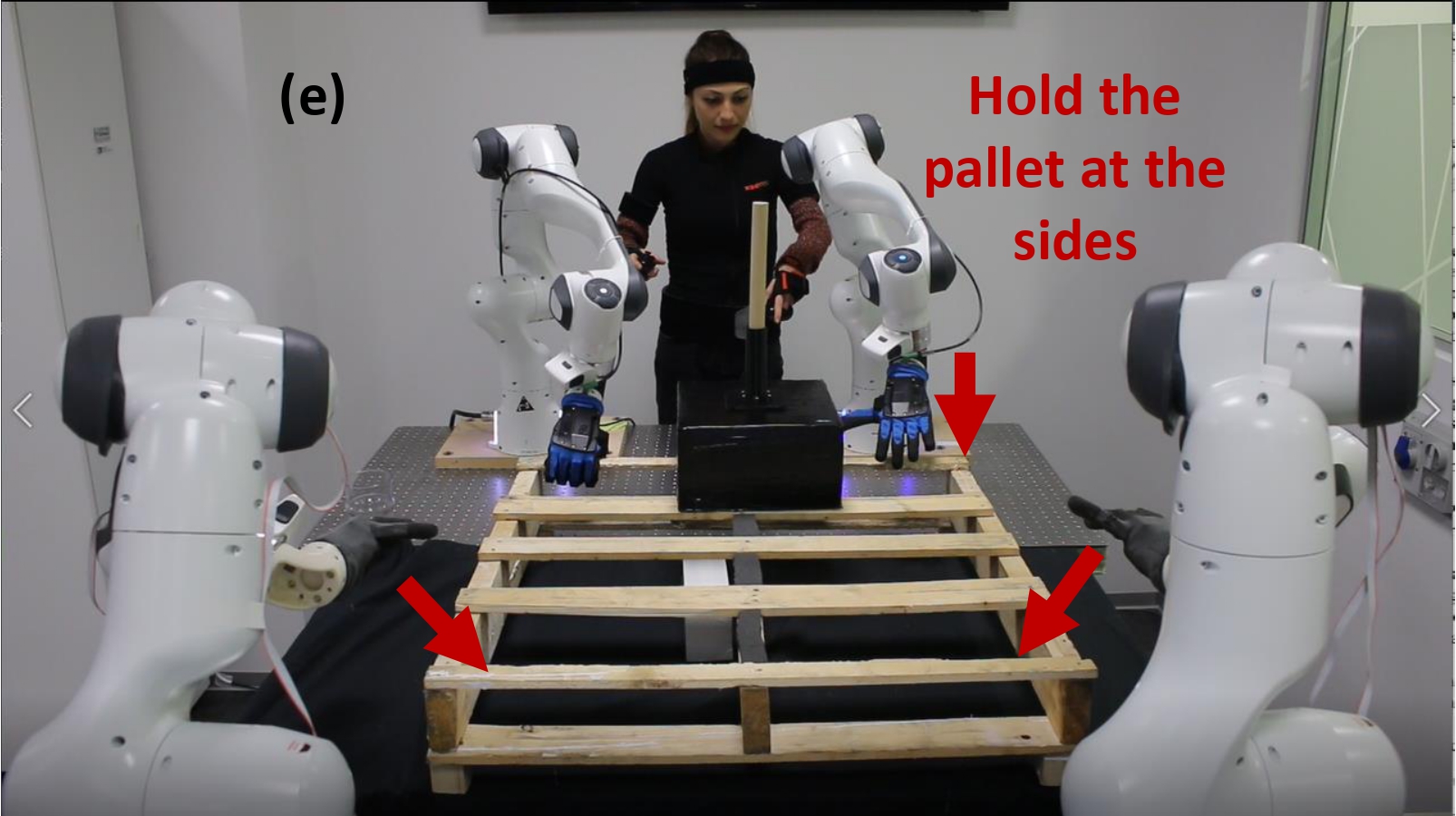}
		\label{fig:task_frame_6}}\hspace{-0.2cm}
	\subfigure{\includegraphics[width = 0.247\linewidth,trim={2.1cm 0cm 2cm 0.5cm},clip]{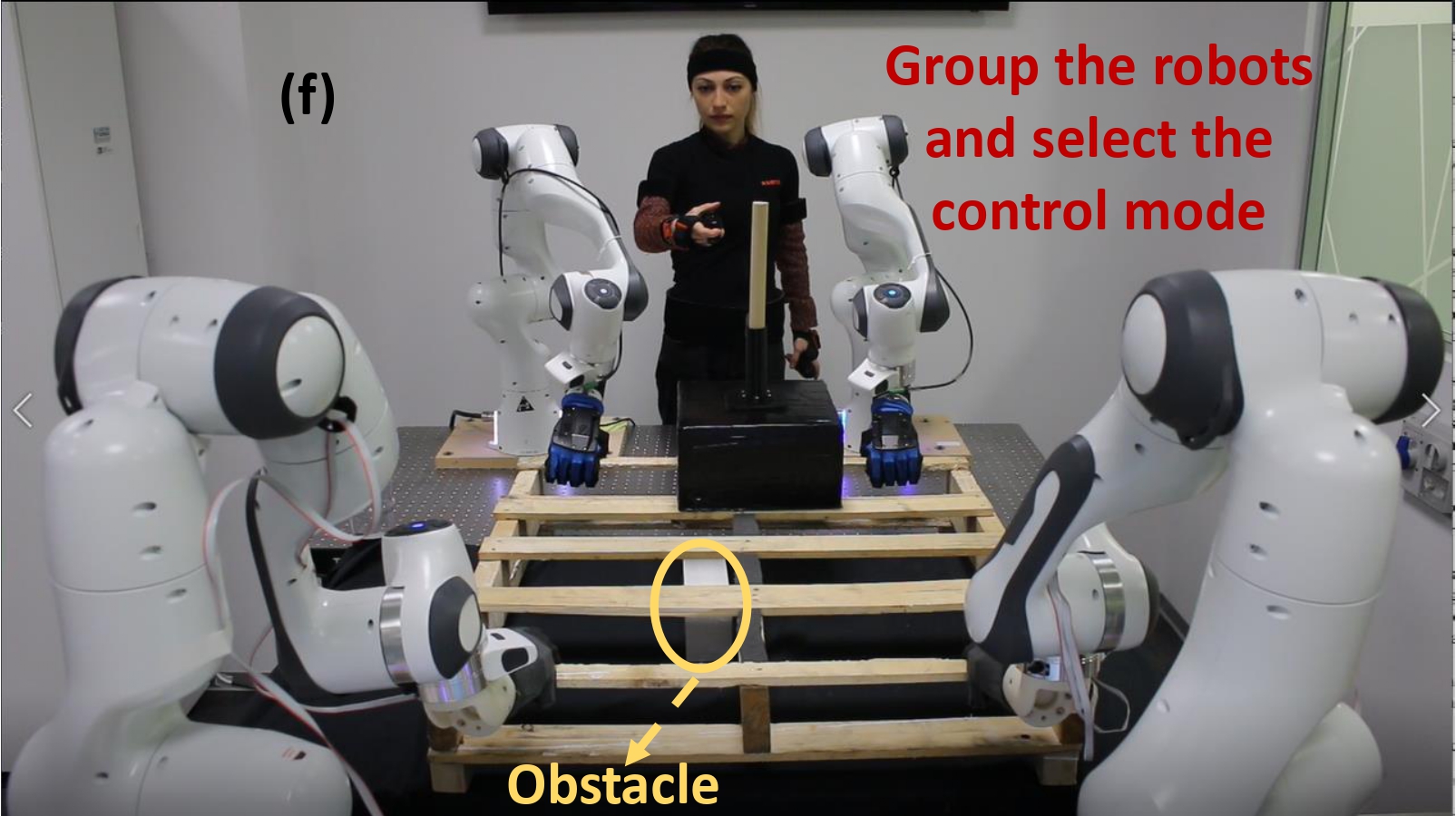}
		\label{fig:task_frame_7}}\hspace{-0.2cm}
	\subfigure{\includegraphics[width = 0.247\linewidth,trim={2.1cm 0cm 2cm 0.5cm},clip]{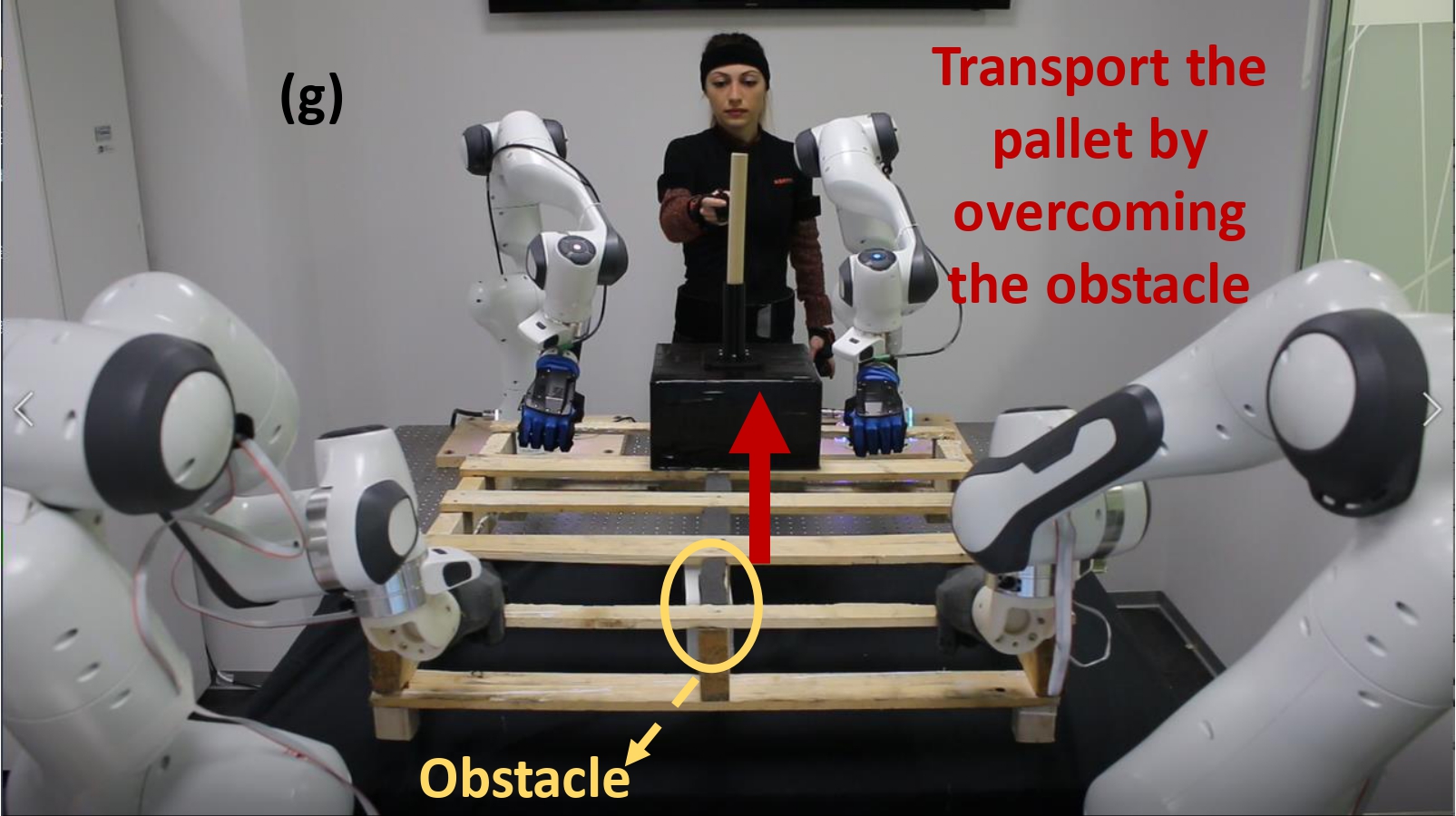}
		\label{fig:task_frame_8}}\hspace{-0.2cm}
	\subfigure{\includegraphics[width = 0.247\linewidth,trim={2.1cm 0.2cm 2cm 0.3cm},clip]{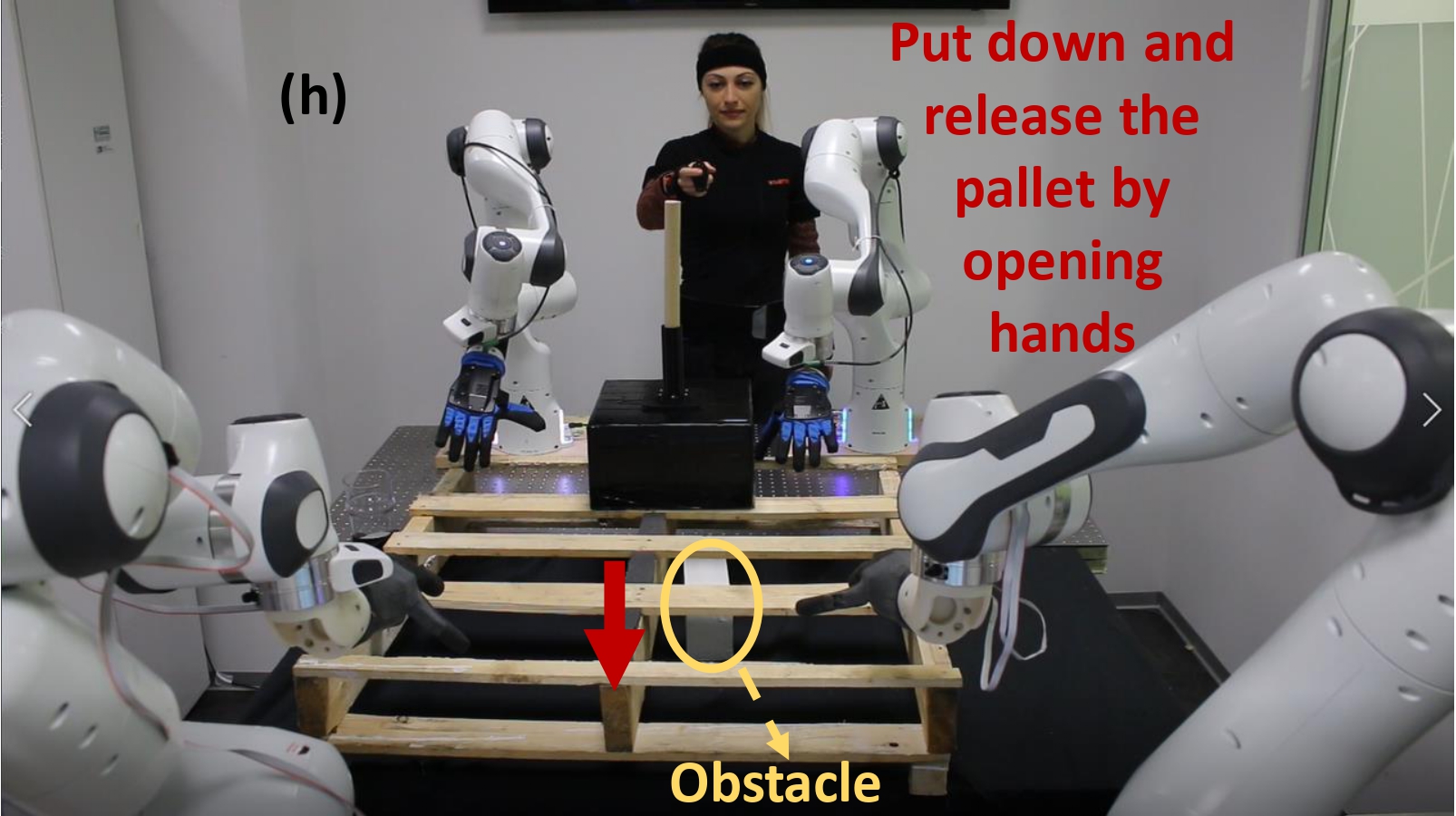}
		\label{fig:task_frame_9}}\hspace{-0.2cm}
		\vspace{-0.3cm}

	\caption{The snapshots of the experiment: (a)-(c) starting, picking, and placing the box and grabbing the peg; (d) executing the peg-in-hole task; (e) grabbing the pallet; and (f)-(h) transporting the pallet (lifted, translated horizontally, and put back without hitting the obstacle). The yellow circles indicate the peg, the box, and the obstacle; the red arrows and the texts show the motion and describe the action respectively.} \vspace{-6mm}
	\label{fig:mixed_task_frame}
\end{figure*}\vspace{-2mm}

\vspace{-3mm}
\subsection{Experimental Procedure}\label{subsec:procedure}
The experiments were conducted with twelve naive healthy volunteers (age: $27.16 \pm 1.58$ years, 6 female) who have not teleoperated the collaborative robot arm before. After explaining the experimental procedure, written informed consent was obtained. The conducted experiment was divided into two sub-tasks, namely 1) Assembly of an object (\ref{subsubsec:subtask_1}) and 2) Manipulation of a bulky pallet (\ref{subsubsec:subtask_2}), which were specifically designed to assess the proposed framework in real-world scenarios.
A possible evolution of the robot poses during the whole experiment is depicted in Fig. \ref{fig:robot_pos} and also shown in the video footage\textsuperscript{\ref{video}} for the demonstrations of the control modalities. The experiment was performed twice by each participant and the trials were not confined within a time limit\footnote{The whole experimental procedure was in accordance with the Declaration of Helsinki, and the protocol was approved by the ethics committee Azienda Sanitaria Locale (ASL) Genovese N.3 (Protocol IIT\_HRII\_ERGOLEAN 156/2020).}.

\subsubsection{Sub-task 1: Assembly of an object}
\label{subsubsec:subtask_1}
the participant picked the peg and the hole (Fig.~\ref{fig:task_frame_2}), purposely placed in opposite sides where only one robot can reach, and brought them to the middle area  (Fig.~\ref{fig:task_frame_2}-\subref{fig:task_frame_4}). The peg used was a cylinder a length of $25$~cm and a diameter of $32$~mm, whereas the hole had an internal diameter of $34$~mm. After bringing them, a classic peg-in-hole task was performed, representing complex interactions with the remote environment. The sub-task is finished when the assembled peg and the hole are placed on the pallet (Fig.~\ref{fig:task_frame_5}).

\subsubsection{Sub-task 2: Manipulation of a bulky pallet}
\label{subsubsec:subtask_2}
the participant must hold the different locations of the pallet by taking control of the robots one by one (Fig.~\ref{fig:task_frame_6}). Subsequently, the pallet had to be lifted, translated horizontally, and put back on the table by employing all the robots (Fig.~\ref{fig:task_frame_7}-\subref{fig:task_frame_9}). In order to constrain the motion of the operator and make the task challenging, we placed an obstacle ($12$~cm tall and $5$~cm width) at the right side of the central wooden board of the pallet (see Fig.~\ref{fig:task_frame_2}). Moreover, the object on the pallet should be carried without falling. The pallet used has dimensions $82\times 78\times 15$~cm and weight of $8.2$~kg. It was intentionally chosen bulky and heavy, so to force the use of all four arms for its transportation.

\vspace{-3mm}
\subsection{User Study}\label{subsection:user_study}
After the experiments, each participant was asked to fill out two 5-points Likert Scale questionnaires: the System Usability Scale (SUS) \cite{SUS} (see Tab. \ref{table:sus}), and a custom questionnaire (see Tab. \ref{table:custom_quest}) considering both trials. While SUS is a general questionnaire evaluating the usability performance of the system in the aspects of effectiveness, efficiency, and overall ease of use, the custom one was formulated to specifically evaluate and address the developed framework. The original SUS scores of the participants were normalized and converted to percentile ranking (between 0-100) and the average of it was calculated and reported in Tab. \ref{table:sus} for the proposed framework.

\begin{figure*}[ht]\vspace{-1mm}
	\centering
	\includegraphics[width=1\linewidth,trim={0cm 0.15cm 0cm 0cm},clip]{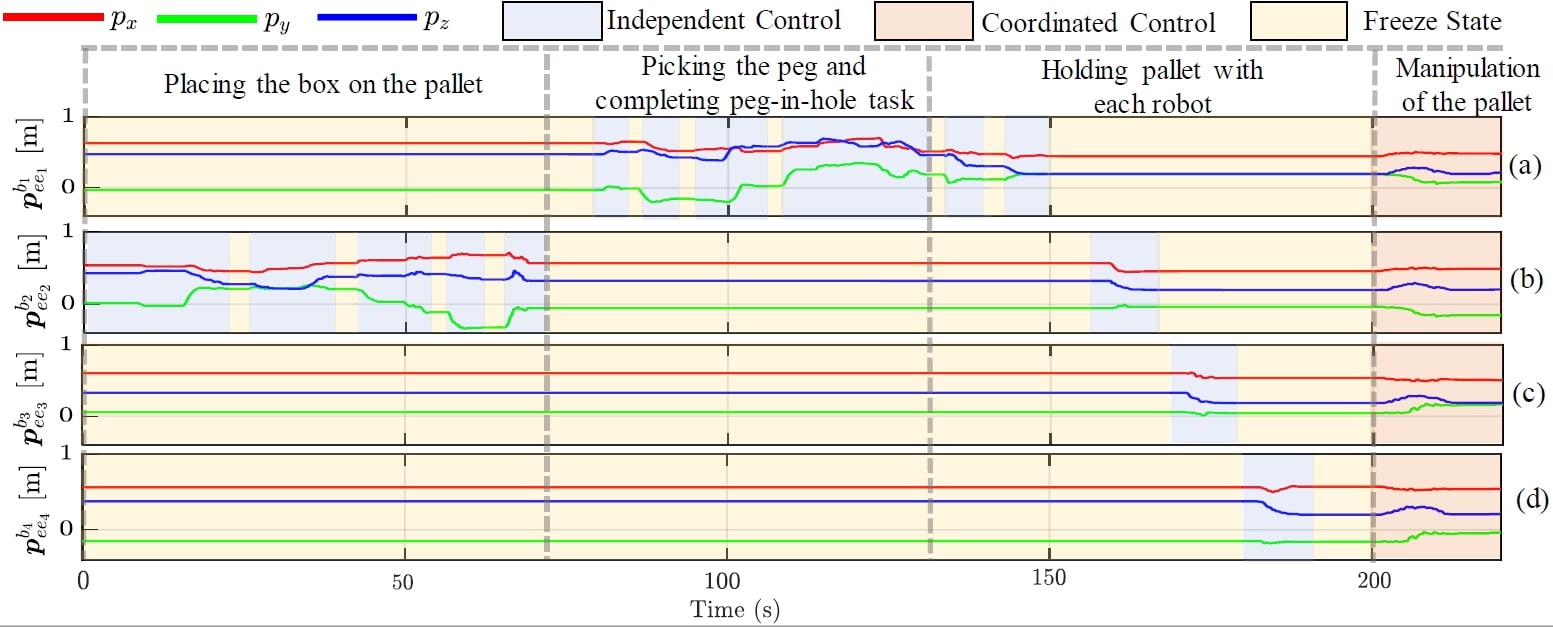} \vspace{-5mm}
	\caption{End-effector positions of the robots for an experimental trial. Light blue, orange and yellow shaded areas indicate the user's preferred control modality and the time range it covers.}\vspace{-5mm}
	\label{fig:robot_pos}
\end{figure*}\vspace{-2mm}

\begin{figure}
	\centering
	\includegraphics[width=1\linewidth]{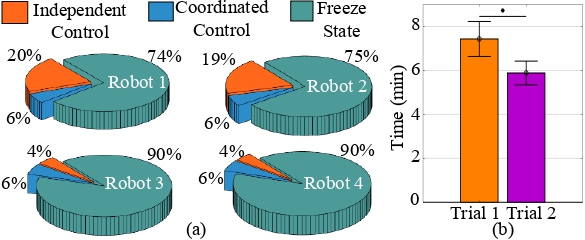}\vspace{-2mm}
	\caption{(a) The average distribution of control modes used by each robot that covers all subjects and trials and (b) the means and the standard errors of the task completion times, *: p $< 0.01$.}	\label{fig:pie_charts_and_time}\vspace{-6mm}
\end{figure}
\vspace{-1mm}

\section{Results and Discussion}\label{sec:results}

The phases of the experiment, end-effector positions of the robots, and used control modalities for an example trial are depicted in Fig.~\ref{fig:robot_pos}.
At first, the user chose the robot on her/his left side to operate with the independent control, since the box was reachable by only that robot and cannot be manipulated by the others (Fig.~\ref{fig:task_frame_2}).
Subsequently, the user selected the \textit{Freeze State} for this robot and started to control the robot on the right side, in order to take the peg and place it inside the hole (Fig.~\ref{fig:task_frame_5}). During this interval, by getting rid of the telemanipulation obligation of the unused robotic arm (entering in the \textit{Freeze State}), the user can be more focused on the execution of the task with the other.
Moreover, the general tendency among the participants was to enter the \textit{Freeze State} while performing the sub-tasks, position their hands in a more advantageous and comfortable position, and continue the task from where it left off (see Fig.~\ref{fig:robot_pos} for an example).

During system reconfiguration, the framework ensures that there were no jumps in robots' positions during transitions by remapping of the human hand to the end-effectors (or virtual frame in CC strategy, see Sec. \ref{subsec:control_strategies}).
Thanks to this feature, the robots, which were previously in \textit{Freeze State}, were brought to the desired point of the pallet one by one and grasped it \added{(Fig.~\ref{fig:task_frame_6})}. \added{All participants carried out the task up to this stage using \textit{Independent Control} and \textit{Freeze State} similar to Fig.~\ref{fig:robot_pos}}. Then, the pallet manipulation was carried out by uniting all the robots in one group and assigning their control to one human arm \added{(Fig.~\ref{fig:task_frame_7}-\subref{fig:task_frame_9}).}
The pallet is transported using \textit{Coordinated Control} of multiple robot arms \added{(22 out of 24 trials across the participants)}, focusing on the motion of the object instead of considering the individual positions of the robots.
Overall, Fig.~\ref{fig:robot_pos} shows each control modality has been used according to the phase of the task as presumed.
In addition, the reconfigurability of the proposed telemanipulation framework gives flexibility to the user to exploit them. \added{As we can see from Fig.~\ref{fig:pie_charts_and_time}(a), the average distributions (for all subjects) of the IC mode were higher than CC mode for the robots 1 and 2 considering they were employed to perform the more dexterous part of the experiment (\textit{Sub-task 1: Assembly of an object}).
The distributions of the IC and CC mode of the robots 3 and 4 were close since after positioning and grasping the pallet, the general tendency among the subjects was to carry it with CC mode.
In average, the participants were able to complete the second trial 1.548 minute faster (20.83\% lower that the first, $p < 0.01$, see Fig.~\ref{fig:pie_charts_and_time}(b)), possibly due to the increased engagement with the framework.}

The results of the SUS questionnaire are reported in the last column of Tab. \ref{table:sus}. The average percentile ranking of the participants (last row of the table) was \added{$81.25 \pm 10.01$}, indicating a high degree of acceptance (between “good” and “excellent” of the overall SUS \cite{AaronSus}). Analyzing the individual question results, we can see that the high scores of positive statements (see Table~\ref{table:sus}) except Q9 implied the system is well-integrated and favorable for the users. The low scores of the negative statements pointed out the framework is not complex, consistent, accommodating and a short familiarization period is sufficient to adapt. This is supported by the decrease in the task completion time (see Fig.~\ref{fig:pie_charts_and_time}(b)) 
However, the scores of participants' self-sufficiency apprehension-related statement Q4 was slightly better than average satisfaction. This is possibly due to the shortcoming of the participants' teleoperation experience, though we can deduce from the overall SUS result that the proposed framework is feasible even for untrained people. 

Moreover, according to the high scores of the custom questionnaire (see Tab. \ref{table:custom_quest}), the users favored the overall framework. Note that, the adaptation to a generalized multi-arm telemanipulation system can be complex for a first-time user. In spite of that, the proposed control modalities together with our reconfigurable control architecture seem to tackle this problem. Additionally, the advantage of the tele-impedance control is demonstrated before in \cite{AjoudaniPegHole} for the peg-in-hole task. We purposely included this task in our experimental scenario since the complex interactions with the remote environment can adversely affect the participants' impressions. The conducted user study infers that the employed tele-impedance paradigm was a constructive complementary for the framework.
\vspace{-1mm}

\newcommand{\len}{1.5cm}

\begin{table}[!h]\vspace{-1mm}
    \centering
    \caption{System Usability Scale (SUS) Questionnaire \cite{SUS} results}\vspace{-2mm}
    \label{table:sus}
    \def\arraystretch{1}
    \begin{tabular}{l l l}
    \hline
    \hspace{-2mm}\textbf{ID}  &\hspace{-2mm}\textbf{Statement} &\textbf{\hspace{-4mm}Raw Score} \\
    \hline
    \hspace{-2mm}Q1 & \hspace{-4mm} I think that I would like to use this system frequently&{\hspace{-3mm}$4.42 \pm 0.66$}\\
    \hspace{-2mm}Q2 & \hspace{-4mm} I found the system unnecessarily complex &{\hspace{-3mm}$1.5 \pm 0.79$} \\
    \hspace{-2mm}Q3 & \hspace{-4mm} I thought the system was easy to use &{\hspace{-3mm}$4.16 \pm 0.71$}  \\
    \hspace{-2mm}Q4 & \hspace{-4mm} I think that I would need the support of a technical &{\hspace{-3mm}$2.16 \pm 0.71$} \\
    &\hspace{-4mm} person to be able to use this system\\
    \hspace{-2mm}Q5 & \hspace{-4mm} I found the various functions in this system were well &{\hspace{-3mm}$4.41 \pm 0.66$}  \\
    &\hspace{-4mm} integrated \\
    \hspace{-2mm}Q6 & \hspace{-4mm} I thought there was too much inconsistency in this system &{\hspace{-3mm}$1.16 \pm 0.38$} \\
    \hspace{-2mm}Q7 & \hspace{-4mm} I would imagine that most people would learn to use this&{\hspace{-3mm}$4.08 \pm 0.66$}  \\
    &\hspace{-4mm} system very quickly\\
    \hspace{-2mm}Q8 & \hspace{-4mm} I found the system very cumbersome to use &{\hspace{-3mm}$1.75 \pm 0.86$}  \\
    \hspace{-2mm}Q9 & \hspace{-4mm} I felt very confident using the system &{\hspace{-3mm}$3.83 \pm 0.71$}  \\
    \hspace{-2mm}Q10 & \hspace{-4mm} I needed to learn a lot of things before I could get going&{\hspace{-3mm}$1.75 \pm 0.75$}  \\
    &\hspace{-4mm} with this system \\
    \hline 
    & \textbf{\hspace{-4mm}Average Percentile Ranking} &\textbf{{\hspace{-5mm}$81.25 \pm 10.01$}} \\
    \hline     
\end{tabular}
\end{table}\vspace{-2mm}

\begin{table}[!h]\vspace{-3mm}
    \centering
    \caption{Custom Questionnaire statements and results}\vspace{-2mm}
    \label{table:custom_quest}
    \def\arraystretch{1}
    \begin{tabular}{l l c}
    \hline
    \hspace{-2mm}\textbf{ID}  &\textbf{\hspace{-3mm}Statement} &\textbf{Score}  \\
    \hline
    \hspace{-2mm}Q1 &\hspace{-4mm} The robots were following my movements well &{$4.27 \pm 0.64$} \\
    \hspace{-2mm}Q2 &\hspace{-4mm} It was easy to create robot groups and control them &{$4.36 \pm 0.80$}  \\
    \hspace{-2mm}Q3 &\hspace{-4mm} I benefited from the system reconfigurability during &{$4.27 \pm 0.64$}  \\
    &\hspace{-4mm} different periods of the task\\
    \hspace{-2mm}Q4 &\hspace{-4mm} I found \emph{Freeze State} beneficial &{$5\pm 0$}   \\
    &\hspace{-4mm} during different periods of the task  \\
    \hspace{-2mm}Q5 &\hspace{-4mm} I found \emph{Coordinated Control Strategy} beneficial &{$4.36 \pm 0.80$}   \\
    &\hspace{-4mm} during different periods of the task  \\
    \hspace{-2mm}Q6 &\hspace{-4mm} I found \emph{Independent Control Strategy} beneficial &{$4.72 \pm 0.46$} \\
    &\hspace{-4mm} during different periods of the task  \\
     \hline     
\end{tabular}
\end{table}\vspace{-5mm}

\section{CONCLUSION}
\label{sec:conclusions}

This paper presents a novel multi-arm telemanipulation framework that provides intuitive control of both individual and the various combination of any number of robotic arms. With our architecture, the human operator can choose the proposed control modalities and the manipulators that make the task convenient to execute. The experiments demonstrated the advantages of having different control modalities and multi-robot control, since the users accomplished to perform a complex task which was impossible with one or two robotic arms. Moreover, the results showed promising applicability of our framework thanks to its flexibility and intuitive reconfigurability.

Future works will focus on increasing usability and user experience on different aspects. First, we'll focus on augmenting user control over the remote interaction. This could be, e.g., providing visual feedback of the contact forces exerted on the manipulated objects (which are now implicitly controlled through the distance regulation between robots, or better to automatically control them, in order to further unburden user duties and increase the task performance. Besides, the joystick interface can be replaced with a more advanced GUI and a voice recognition system to reconfigure the control architecture. This may also increase the variety of the user inputs and enrich control over the system.  
\vspace{-4mm}

\bibliographystyle{IEEEtran}
\bibliography{IEEEabrv,biblio}

\end{document}